\documentclass[12pt,twoside,a4paper,reqno]{amsart}
\usepackage{bulletinUPB}
\usepackage{graphics}
\usepackage{amssymb}
\usepackage{eucal}
\usepackage{amsmath}
\usepackage{color}
\usepackage{multirow} 
\usepackage{graphicx} 

\usepackage[style=numeric, giveninits=true, backend=biber, maxnames=999, minnames=1,url=false,natbib=true]{biblatex}
\usepackage{microtype}

\defbibheading{bibliography}[\refname]{%
  \section*{\centering\normalfont{\textls{\MakeUppercase{References}}}}}

\usepackage[hidelinks]{hyperref}
\info{C}{85}{4}{2023}
\title{Deepfake Sentry: Harnessing Ensemble Intelligence for Resilient Detection and Generalisation}
\address{$^*$ These authors contributed equally to this work.}
\author{L.-D. \c{S}tefan}
\address{$^1$Research Scientist, AI Multimedia Lab, Univ. ``Politehnica" of Bucharest, Romania, e-mail: {\tt liviu\_daniel.stefan@upb.ro.}}
\author{D.-C. Stanciu}
\address{$^2$Engineer, AI Multimedia Lab, Univ.``Politehnica" of Bucharest, Romania.}
\author{M. Dogariu.}
\address{$^3$Lecturer, Dept. of Applied Electronics and Information Engineering, Univ.``Politeh\-nica" of Bucharest, Romania.}
\author{M.G. Constantin, A.C. Jitaru, and B. Ionescu}
\address{$^4$Professor, Dept. of Applied Electronics and Information Engineering, Univ. ``Po\-li\-teh\-ni\-ca" of Bucharest, Romania.}

\begin{document}
\pagestyle{headings}
\maketitle

\begin{abstract}

Recent advancements in Generative Adversarial Networks (GANs) have enabled photorealistic image generation with high quality. However, the malicious use of such generated media has raised concerns regarding visual misinformation. Although deepfake detection research has demonstrated high accuracy, it is vulnerable to advances in generation techniques and adversarial iterations on detection countermeasures. To address this, we propose a proactive and sustainable deepfake training augmentation solution that introduces artificial fingerprints into models. We achieve this by employing an ensemble learning approach that incorporates a pool of autoencoders that mimic the effect of the artefacts introduced by the deepfake generator models. Experiments on three datasets reveal that our proposed ensemble autoencoder-based data augmentation learning approach offers improvements in terms of generalisation, resistance against basic data perturbations such as noise, blurring, sharpness enhancement, and affine transforms, resilience to commonly used lossy compression algorithms such as JPEG, and enhanced resistance against adversarial attacks.

\end{abstract}

\begin{Keywords}
Deepfake detection, deepfake generalisation, video or image manipulation, digital media forensics, perturbation attacks.
\end{Keywords}

\section{Introduction}

Modern access to the Internet has surged in recent years, with virtually everybody being connected to some sort of media, be it social websites, news outlets, entertainment platforms or telecommunications services. This large influx of users also lead to an increase in the amount of digital information available online. Consequently, the quality, correctness and truthfulness of such information gains increasing importance, especially since most people are unaware of the impact that false or erroneous information can have on the public opinion.

One large-scale statistic concerning Internet users' behaviour comes from The European Commission, which has published recent data on Information and Communication Technology users showing that only a mere 35.25\% of the surveyed individuals that used the Internet in the last 3 months have checked the truthfulness of the information or content they found on the internet news sites or social media~\cite{eurostat2021ict}. This is an astoundingly low number given the unrestricted access to information that the citizens of the European Union have. Even more so, younger generations tend to be more inclined to checking online information as opposed to older ones, thus leaving a large amount of the European population at risk of being deceived by disinformation under digital form.

In the context of rapid AI development, a new threat appeared in the shape of fake images or videos generated by deep learning algorithms, a technique coined as "deepfake". This is usually represented by visual manipulation of people's identities, leading to fake content involving these persons with potential negative impact on their reputation. The preferred way of achieving this is through deep generative models, most notably Generative Adversarial Networks (GANs)~\cite{NIPS2014_5ca3e9b1}. Even though there are several attempts at using this technique as an entertainment tool such as DeepFaceLab~\cite{perov2020deepfacelab}, DeepSwap~\cite{deepswap}, FaceSwap~\cite{FaceSwapDevs} or FaceApp~\cite{faceapp}, deepfakes remain a notable threat to public safety, especially since it can be used for blackmail, mischievous impersonation of other people or defamation.

In response to deepfakes' negative effect, researchers started paying more attention to detecting malevolent interference with digital imagery as a forensic method. It is customary for the deepfake algorithms to focus on one of the two aspects: either manipulating a person's facial expression or manipulating a person's entire identity~\cite{rossler2019faceforensics}. Therefore, deepfake detectors focus on artifacts introduced by these types of operations. In a simplified form, deepfake detectors act as binary classifiers, being required to tell whether an image or a video sequence has been altered (is fake) or not.

However, visual content is frequently impacted by various sources of noise, especially in the case of in-the-wild contexts, which can further hinder the decision-making process of the detector. In this paper we investigate the impact of a range of perturbations on the performance of a deepfake detector, and examines the individual effects of these perturbations through experimental analysis on three widely used datasets: FF++~\cite{rossler2019faceforensics}, DFDC Preview~\cite{dolhansky2019deepfake} and Celeb-DF~\cite{li2020celeb}. Furthermore, we propose an autoencoder ensemble-based approach to alleviate the unwanted effect of added perturbations. We perform tests on 2 types of augmentation --- we augment the training datasets 1) with carefully tuned transformations of the input samples and 2) with an ensemble of autoencoders specifically designed to bring deepfake-like artefacts in the training dataset, so as to offer the detector more positive samples to train on. Extensive experiments prove that our methods improve the AUC score in all cases on a general deepfake detector such as Xception~\cite{chollet2017xception}. A key advantage of our proposed approach is its model-agnostic nature, which allows it to be applied to any deepfake detector. This is due to the fact that our method acts solely on the training data, without taking the specific detector model into account. The contributions of our work are threefold:
\begin{itemize}
    \item We introduce 2 types of model-agnostic augmentations that improve the affected detector's performance.
    \item We proved our proposed approach's robustness by performing rigorous tests on 3 different popular datasets.
    \item We test deepfake detectors against 9 different types of perturbations, 5 different levels of lossy compression and white-box and black-box adversarial attacks.
\end{itemize}

The rest of this paper is organized as follows: we present the current state-of-the-art in Section~\ref{sec:rel_works}, we describe our proposed ensembling approach using autoencoders in Section~\ref{sec:approach}, in Section~\ref{sec:attacks} we describe the types of attacks and perturbations that we applied. Section \ref{sec:experiments} presents our experimental setup, details regarding training, datasets and evaluation metrics, while Section \ref{sec:results} presents the results of those experiments and discussions highlighting the quantitative improvements of our proposed method, in multiple real-life scenarios. Finally, Section \ref{sec:conclusions} presents a summarised view of the paper and of the benefits of using our proposed method.

\section{Related Works}
\label{sec:rel_works}

The objective of \textit{deepfakes detection} is to categorise the input media into either authentic or artificial. There exist noteworthy ongoing endeavours aimed at the development of forensic methodologies with the purpose of discerning synthesized or manipulated image and video recordings. These methodologies can be broadly classified into two overarching categories: high-level and low-level.

\textit{1. High-level deepfake detection}.
High-level forensic techniques place emphasis on semantically meaningful attributes, encompassing aspects such as incongruities in eye blinks, head-pose, physiological signals, and distinctive behavioural patterns. For example, \cite{li2018ictu} aim to identify the absence of inherent physiological signals in synthesized videos that are typically inadequately captured. These signals encompass spontaneous and involuntary physiological manifestations, including respiration, pulse, and ocular movements, which are frequently overlooked during the fabrication of counterfeit videos. \cite{9141516} make the observation that deep learning-based detectors exhibit limited efficacy in detecting counterfeit content, owing to the impressive realism achieved by generative models. They put forth the central argument that biological signals embedded within portrait videos can serve as an implicit indicator of authenticity, as these signals are neither spatially nor temporally preserved in fabricated content. \cite{li2019exposing} present a deep learning-based approach which offers a proficient means of discriminating DeepFake videos by leveraging a notable characteristic exhibited by such videos. Specifically, owing to computational constraints and time limitations inherent in the DeepFake algorithm, the synthesis of face images is confined to a predetermined size, necessitating an affine warping process to align them with the facial configuration of the source material. Consequently, this warping procedure imparts perceptible artefacts stemming from the resolution disparities between the warped facial region and its surrounding context. \cite{9413139} investigate the synthesis mechanism employed in deepfake videos as well as the distinctive nature of human eye movements. They examine the patterns of eye movements and utilize this information to unveil deepfake videos. The authors establish four features related to binocular movements, which are subsequently inputted into a classification algorithm. \cite{Agarwal_2019_CVPR_Workshops} demonstrate that individuals display distinct patterns of facial and head movements during speech. Furthermore, they observe that the generation of face-swap, lip-sync, and puppet-master deep fakes tends to disrupt these patterns due to the control exerted by an impersonator (in the case of face-swap and puppet-master) or the decoupling of the mouth from the rest of the face (in the case of lip-sync). Leveraging these consistent patterns, the authors construct soft biometric models for prominent individuals and employ these models to discriminate between genuine and fabricated videos.

\textit{2. Low-level deepfake detection}.
Conversely, low-level forensic techniques are designed to identify pixel-level irregularities introduced during the synthesis process. \cite{Yu_2019_ICCV} investigate GAN fingerprints for the purpose of image attribution. They demonstrate that even minor variations in GAN training, such as differences in initialization, can result in distinctive fingerprints that persist across all generated images. Furthermore, the authors uncover the persistent nature of these fingerprints across various frequencies and patch sizes, revealing their independence from GAN artifacts.
\cite{NEURIPS2022_1d051fb6} propose the Local-Temporal-aware Transformer-based Deepfake Detection framework, which introduces a novel methodology for detecting deepfakes by employing Transformers to model patch sequences. This framework places particular emphasis on capturing the temporal consistency within sequences of restricted spatial regions. To achieve this, shallow layers of learned 3D filters are utilized to hierarchically enhance the low-level information in the input data. Additional approaches focus on investigating clues in the frequency domain. For instance, \cite{10.1007/978-3-030-58610-2_6, li2021frequency} incorporate low-level frequency pattern learning into convolutional neural networks (CNNs) by employing the Discrete Cosine Transform (DCT) transform. This integration aims to enhance the generalizability of the models. Conversely, \cite{liu2021spatial} undertake a theoretical analysis, highlighting the increased sensitivity of phase information to upsampling. Consequently, they assert the significance of such low-level features, which surpass the importance of high-level semantic information. We can notice that the inherent advantage of low-level approaches lies in their capability to detect artefacts that may elude visual detection. Nevertheless, in contrast to high-level techniques, they encounter challenges in terms of generalisation to novel datasets and can be vulnerable to the influence of laundering procedures, such as compressions or geometric transformations.

While image forensics models have achieved high levels of proficiency in discriminating synthetic images from authentic ones, these models remain vulnerable to simple perturbations via \textit{deepfake disruptions}. DeepFake disruption involves the proactive introduction of diverse perturbations into the source image to create imperceptibly altered images, thereby aiming to disrupt the process of DeepFaking. \cite{9533868} employed differentiable random image transformations to generate adversarial faces that exhibit discernible artifacts. \cite{huang2021initiative} propose a two-stage training framework for the development of a "poison" generator intended for face attribute editing and face reenactment deepfakes. This work aim to inject perturbations that generate the desired "venom" via surrogate models. \cite{ijcai2022p107} conduct an investigation into the susceptibility of GANs in image synthesis and aim to identify a resilient form of adversarial perturbation capable of withstanding significant input transformations. The perturbations are deliberately applied within the Lab color space, with specific attention directed towards the decorrelated a and b channels. By operating within this specific color space, the proposed approach endeavours to alleviate the discernible distortions that were previously observed when employing the RGB color space, even with minor deviations. Instead, the method focuses on generating facial images that manifest a perceptually natural appearance by concentrating on a range of colours that align with semantic categories, thereby ensuring a heightened sense of authenticity.  

Similarly, in this paper, we demonstrate the inherent susceptibility of such forensic classifiers to a multitude of attacks that effectively undermine their predictive capabilities.

\begin{figure*}[!htb]
\centering
    \includegraphics[width = 1\linewidth]{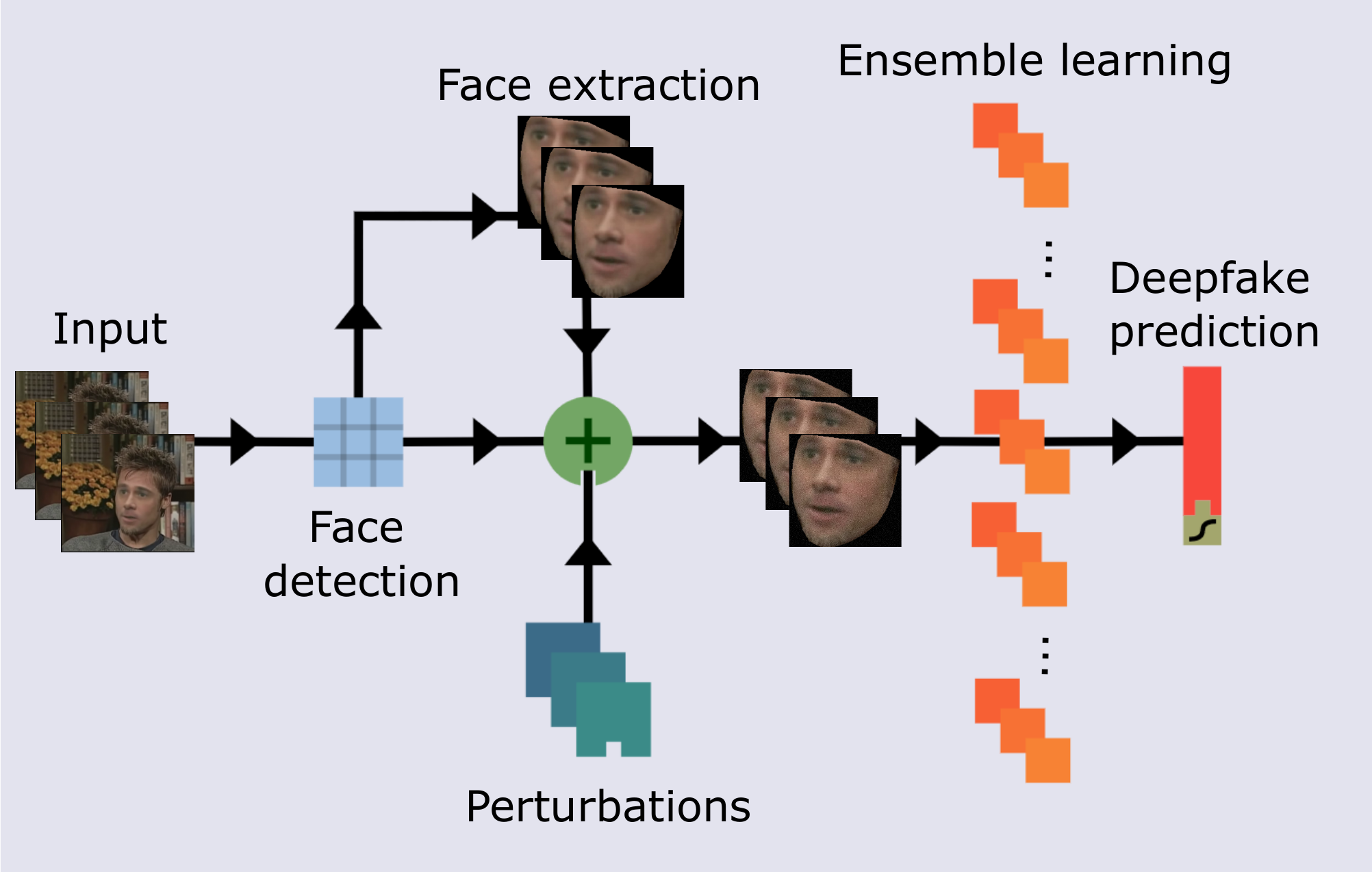}
\caption{The figure illustrates the proposed training paradigm for our deepfake detection model, which comprises of four stages, namely (1) face detection and facial region extraction, (2) insertion of classic perturbations, (3) insertion of artificial fingerprints via an autoencoder ensemble, and (4) deepfake prediction model.}
\label{fig:methodgraph}
\end{figure*}

\section{Approach}
\label{sec:approach}
In this section, we provide the formulation of our proposed deepfake neural ensemble architecture. The overall framework is visualized in Fig.~\ref{fig:methodgraph}. Our proposed training paradigm for deepfake detection is composed of several steps. The first step involves detecting the face regions in the input RGB images and extracting the corresponding facial regions. In the second step, we apply a set of classic perturbations, such as noise, blur, and affine transformations, to the facial regions. The third step involves generating artificial fingerprints using an ensemble of autoencoders, and inserting them into the perturbed facial regions, methodology described in Section~\ref{sec:ensemble}. Subsequently, we ascertain the transferability of these fingerprints from the training data, enabling us to detect deepfakes by classifying images exhibiting similar fingerprints. These artificial fingerprints aim to increase the robustness of the model against adversarial attacks and to improve its generalisation capability. Finally, the perturbed and fingerprinted facial regions are fed into a prediction model, which is responsible for detecting the presence of deepfakes.

Our proposed methodology is characterized by its simplicity and efficacy. This simple methodology ensures that our solution remains agnostic and readily adaptable to a wide range of models, thereby facilitating seamless integration and deployment.

\subsection{Neural Ensemble Architecture for Deepfakes}
\label{sec:ensemble}
Whilst the prevailing body of research on deepfake detection showcases impressive accuracy, it remains susceptible to advancements in generative techniques and adversarial iterations aimed at circumventing detection countermeasures. Every GAN model encompasses numerous parameters, including the distribution of the training dataset, the network architecture, loss formulation, optimization strategy and hyperparameter configurations. Due to the non-convex nature of the objective function and the inherent instability of adversarial equilibrium between the generator and discriminator within GANs, the weights of the model are highly sensitive to their random initializations and do not converge to identical values during training. This implies that while two adequately trained GAN models may exhibit comparable performance, they generate high-quality images in distinct manners. Consequently, the existence and uniqueness of GAN fingerprints can be inferred from these observations. Therefore, we endeavour to develop a proactive and sustainable solution for deepfake detection that remains impervious to the evolution of generative models. Our proposed approach involves introducing artificial fingerprints into the models to achieve this objective.

Let $g: Z \rightarrow X$ be a representation of an image generator. The generator takes a vector in a latent space $Z$ as input and produces a colour image with a predefined resolution. Let $F = \{f_1, f_2, ..., f_k\}$ denote a set of $k$ image-forensic autoencoders. Each autoencoder, $f_i: X \rightarrow X$, takes an image $x\in X$ as input and produces a reconstructed image $f_i(x)$ as output. The ensemble learning involves sequentially passing the original image $x$ through each autoencoder in the set $F$ to generate a chain of reconstructed images: $x_1 = f_1(x), x_2 = f_2(x_1), ..., x_k = f_k(x_{k-1})$. Let $c: X \rightarrow R$ denote an image-forensic classifier. The classifier takes a reconstructed image, $x' \in X$, as input and generates a real-valued scalar output. Higher values of the output indicate an increased probability that the input image is fake or synthetically-generated.

To begin, we establish a set of principles for constructing network architectures, which encompass the following factors: (i) the loss function - Mean Absolute Error or Mean Squared Error, (ii) the upsampling type - Bilinear, Bicubic, Convolutional Transpose, Nearest, (iii) model architecture - number of layers, ranging from 6 layers up to 12 layers, and (iv) layers parameters - kernel sizes, ranging from $3 \times3 $ to $7 \times 7$. Therefore, the network builder is designed to test out rules and search for structures in order of increasing complexity. It is imperative that the resulting output images exhibit dissimilarities from the original images in some manner, thus mimicking the effect of the artefacts introduced by the deepfake generator models. Consequently, these models undergo short training cycles, and the training process ceases if the Mean Absolute Error falls below a predetermined threshold $T$ ($0.03 \leq T \leq 0.25$). This ensures that the autoencoder models introduce a form of variance to the images. The goal is to allow each autoencoder in the pool may introduce its own set of biases or distortions during the reconstruction process. These biases can manifest as subtle artifacts, distortions, or modifications specific to the operations and constraints of each autoencoder. As a result, the reconstructed image may contain residual traces or fingerprints associated with the characteristics of each autoencoder.

\section{Attacks against DeepFake Classifiers}
\label{sec:attacks}

The primary objective of this paper is to investigate deepfakes in real-life scenarios and assess the robustness of image-forensic classifiers against potential attacks. The image-forensic classifier $c$ is a function that maps an input image $x$ to a binary decision output, indicating whether the image is genuine or manipulated. To achieve this, we consider an attack function $A: X \rightarrow X$ that takes a color image as input and outputs an image that is designed to deceive or manipulate the decision-making process of the image-forensic classifier $c$. By subjecting the image-forensic classifier to the attack $A$, we obtain a manipulated image $c(A(x))$ that is intended to evade or confound the classifier, leading to erroneous or unreliable predictions about the authenticity or origin of the image.

We examine the resilience of forensic classifiers by conducting a comprehensive analysis involving various attack scenarios. In these scenarios, we consider two distinct levels of access to the classifier's parameters: (1) white-box attacks, where complete access to the parameters is assumed, and (2) black-box attacks, where no access to the parameters is available. Through these investigations, we aim to evaluate the robustness of the forensic classifiers under different levels of information availability, shedding light on their vulnerability to spatial or frequency domain manipulation methods. Furthermore, we also considered two types of image alterations that mimic real-life situations: basic perturbations and JPEG lossy compression.

\section{Experiments}
\label{sec:experiments}

This section outlines the experimental setup used in this study, which includes the model generation settings that prioritize generalisation and robustness in real-world applications, the datasets used for evaluation and the evaluation metrics.

\subsection{Experimental Setup}

As our primary focus is on the impact of the autoencoder-based augmentation algorithm on real-world applications, we designed the experimental setup with the following considerations: (i) we employed basic, state-of-the-art models, as our focus is on the autoencoder-aided augmented training rather than the model architecture itself; (ii) the models were trained with generalisation in mind to be as model-agnostic as possible, delivering consistent performance irrespective of the employed deepfaking algorithm; and (iii) we trained the autoencoders to realistically replicate images from the original dataset, introducing footprints that are not recognizable with the naked eye.

In our models, we have employed a frame-level approach. Specifically, we have opted to finetune the XceptionNet network~\cite{chollet2017xception} on the FF++ dataset with the objective of producing a binary detection of deepfake/real images. We further test the generalisation capability of the network on the CelebDF and DFDC preview datasets. As input, we utilize images that have dimensions of $3\times 299 \times 299$. Notably, these images exclusively encompass the facial regions present within the frames, while the background regions are deliberately excluded. To identify the coordinates of the facial region, we rely on the facial point coordinates obtained through the utilization of the open-source software OpenFace2~\cite{baltrusaitis2018openface}.

We employ a total of 80 autoencoder models derived from two architectures, a convolutional autoencoder architecture and a U-Net model, following the protocol presented in Section~\ref{sec:ensemble}. The training images are subjected to augmentation techniques including basic perturbations, namely adjustments to sharpness, contrast, perspective, affine transformation, blurring, rotation, color jitter, and noise. Furthermore, different JPEG lossy compression were applied using five compression factors (10, 20, 30, 50, and 80). These perturbations are applied to the images subsequent to their inference through an autoencoder.

The deepfake models were trained using the Adam optimizer with a learning rate of 10\textsuperscript{-4}. To prevent overfitting, a dropout rate of 0.25 was applied before the last fully connected layer, and a weight decay of 0.0003 was used for regularization purposes. 

The pool of autoencoders were trained for a maximum of 50 epochs. During the training process, each batch of images was processed by a network builder, which proposed an autoencoder architecture to introduce deep learning fingerprints.

The adversarial attack was carried out with an epsilon value of 8/255, an alpha value of 2/225, and the number of attack steps set to 10, where the epsilon value determines the maximum perturbation allowed in the input image, the alpha value determines the step size for each iteration of the attack and the number of attack steps determines the number of times the attack will be applied to the input images.

\subsection{Datasets}

\textbf{FaceForensics++ (FF++)}~\cite{rossler2019faceforensics} stands as the preeminent benchmark for deepfake detection, having gained significant popularity. Its unprocessed iteration comprises of a set of 700 videos, designed for rigorous testing. This dataset encompasses 140 authentic videos, alongside 560 manipulated videos generated by four distinct algorithms: Deepfakes (DF)~\cite{FaceSwapDevs}, Face2Face (F2F)~\cite{thies2016face2face}, FaceSwap (FS)~\cite{Kowalski2018}, and NeuralTextures (NT)~\cite{thies2019deferred}.

\textbf{DeepFake detection challenge preview dataset (DFDC Preview})\\~\cite{dolhansky2019deepfake} encompasses a total of 4,113 DeepFake videos, derived from an original collection of 1,131 videos featuring 66 consented individuals. These individuals represent diverse demographics, including various genders, ages, and ethnic groups. The specifics regarding the synthesis algorithm remain undisclosed.

\textbf{Celeb-DF dataset}~\cite{li2020celeb} comprises a total of 590 authentic videos and 5,639 DeepFake videos, constituting a vast collection of over two million video frames. The videos in this dataset have an average duration of approximately 13 seconds and adhere to a standard frame rate of 30 frames per second. The selection of real videos is sourced from publicly accessible YouTube videos, specifically encompassing interviews of 59 renowned personalities. These individuals exhibit a diverse representation across genders, ages, and ethnic groups.

\subsection{Evaluation Metrics}
We evaluate the overall detection performance by assessing the area under the receiver operating characteristic curve (AUC) score at the frame level over a span of 16 consecutive frames. Subsequently, the frame-level output classification scores are averaged to derive the video-level decision. 

\begin{table}[!ht]
\centering
\caption{Deepfake detection results under varios perturbations expressed in AUC on the FF++, CelebDF and DFDC preview datasets where: BL--baseline model, CA--classic augmentation model, EA--ensemble autoencoder augmentation model (proposed), EA+CA--ensemble autoencoder augmentation model combined with classical augmentations (proposed).}
\label{table:deepfake_perturbations}
\renewcommand{\arraystretch}{1.2}
\resizebox{\linewidth}{!}{%
\begin{tabular}{|l|cccc|cccc|cccc|}
\hline
\multirow{2}{*}{\textbf{Perturbation}} & \multicolumn{4}{c|}{\textbf{FF++}} & \multicolumn{4}{c|}{\textbf{CelebDF (generalisation)}} & \multicolumn{4}{c|}{\textbf{DFDC (generalisation)}} \\
& BL & CA & EA & EA+CA & BL & CA & EA & EA+CA & BL & CA & EA & EA+CA \\
\hline

{No distortion} & 0.995 & 0.996 & 0.997 & \textbf{0.999} & 0.748 & 0.770 & 0.805 & \textbf{0.826} & 0.697 & 0.709 & 0.715 & \textbf{0.722} \\
\hline
Adjust Sharpness & 0.969 & 0.991 & 0.993 & \textbf{0.995} & 0.728 & 0.731 & 0.767 & \textbf{0.769} & 0.691 & 0.699 & \textbf{0.713} & \textbf{0.713} \\
Autocontrast & 0.968 & 0.990 & 0.993 & \textbf{0.994} & 0.650 & 0.654 & 0.691 & \textbf{0.701} & 0.698 & 0.701 & 0.703 & \textbf{0.714} \\
Random Perspective & 0.930 & 0.963 & 0.972 & \textbf{0.989} & 0.691 & 0.697 & 0.720 & \textbf{0.722} & 0.670 & 0.670 & \textbf{0.709} & \textbf{0.709} \\
Color Jitter & 0.961 & 0.986 & 0.989 & \textbf{0.994} & 0.741 & 0.764 & 0.796 & \textbf{0.799} & 0.670 & 0.680 & 0.699 & \textbf{0.708} \\
Random Resized Crop & 0.860 & 0.967 & 0.976 & \textbf{0.978} & 0.681 & 0.647 & 0.692 & \textbf{0.713} & 0.646 & 0.628 & 0.688 & \textbf{0.701} \\
Gaussian Blur & 0.962 & 0.987 & 0.990 & \textbf{0.998} & 0.765 & 0.778 & \textbf{0.826} & 0.824 & 0.682 & 0.676 & \textbf{0.716} & 0.713 \\
Random Noise & 0.975 & 0.984 & 0.991 & \textbf{0.995} & 0.731 & 0.764 & 0.803 & \textbf{0.823} & 0.661 & 0.682 & 0.692 & \textbf{0.704} \\
Random Rotation & 0.956 & 0.983 & 0.985 & \textbf{0.992} & 0.745 & 0.765 & 0.783 & \textbf{0.785} & 0.665 & 0.671 & 0.698 & \textbf{0.710} \\
Random Affine (A)  & 0.810 & 0.844 & 0.860 & \textbf{0.906} & 0.612 & 0.620 & 0.658 & \textbf{0.698} & 0.652   & 0.631   & 0.656   & \textbf{0.676}  \\
Random Affine (B) & 0.915 & 0.949 & 0.959 & \textbf{0.966} & 0.697 & \textbf{0.731} & 0.716 & 0.726 & 0.688   & 0.689   & 0.673   & \textbf{0.699} \\
\hline
Average & 0.936 & 0.967 & 0.973 & \textbf{0.982} & 0.708 & 0.720 & 0.750 & \textbf{0.762} &  0.674 & 0.676 & 0.696 & \textbf{0.706} \\

\hline
\end{tabular}
}
\end{table}

\section{Results and Discussions}
\label{sec:results}

In this section, we present extensive experimental results to demonstrate the efficacy of our proposed approach. In this regard, we have trained four different deepfake classifier models with the following descriptions: (i) \textbf{Baseline Model (BL)}: This model is trained solely on the original data, without any data augmentation applied, (ii) \textbf{Classic Augmentation Model (CA)}: This model is trained using classical image augmentation techniques, such as Color Jitter, Random Noise, Gaussian Blur, etc., and different JPEG compression levels, (iii) \textbf{Ensembled Augmentation Model (EA)}:  This model is trained using an ensemble of autoencoders, and (iv) \textbf{Ensembled Augmentation + Clasical Augmentation Model (EA+CA)}: This model combines classical image augmentation techniques with an ensemble of autoencoders. All the models are trained on the FF++ dataset. We further test the generalisation capabilities of the models on the CelebDF and DFDC preview datasets.

Table~\ref{table:deepfake_perturbations} presents the performance of our proposed models when confronted with various forms of perturbations, as measured on the FF++, CelebDF, and DFDC preview datasets. The ability to withstand these types of perturbations is critical as they may be encountered in real-world image processing scenarios, and thus, evaluating the model's robustness is of utmost importance. 

First, it is important to examine the baseline performance, with no distortion (BL). This model achieves an almost perfect AUC on FF++. However, the performance does not translate to other datasets, indicating poor generalisation. Generally, all models (CA, EA, EA+CA) show an increase in performance as we move from the "No distortion" to the subsequent perturbation types. This suggests that perturbations can help improve deepfake detection, and the proposed ensemble autoencoder augmentation model (EA) independently, or combined with classical augmentations (EA+CA) consistently outperforms the baseline model (BL) and the classic augmentation model (CA) across all perturbations. 

On average, the CA, EA and EA+CA models showed an improvement in AUC over BL of 3.3\%, 3.9\% and 4.9\%, respectively, on FF++. We can also observe the generalisation results on the CelebDF and DFDC preview datasets, which are unknown to the detector, making the task closer to real-life scenarios, where we are confronted with new data, unseen in previous training iterations. These are also more challenging datasets than FF++, therefore the presented results are more insightful and hold more relevance. On average, the CA model showed an improvement in AUC of 1.7\% on CelebDF, and 0.3\% on DFDC preview, while the EA model showed an improvement of 5.9\% on CelebDF, and 3.3\% on DFDC preview. The EA+CA model showed the greatest improvement, with 7.6\% on CelebDF, and 4.7\% on DFDC preview. The presented results prove that our proposed model improves generalisation, as CelebDF and DFDC Preview were not used for training.

\begin{table}[!ht]
\centering
\caption{Deepfake detection results under varios JPEG compression levels expressed in AUC on the FF++, CelebDF and DFDC preview datasets where: BL--baseline model, CA--classic augmentation model, EA--ensemble autoencoder augmentation model (proposed), EA+CA--ensemble autoencoder augmentation model combined with classical augmentations (proposed).}
\label{table:deepfake_jpeg}
\renewcommand{\arraystretch}{1.2}
\resizebox{\linewidth}{!}{%
\begin{tabular}{|l|cccc|cccc|cccc|}
\hline
\multirow{2}{*}{\textbf{JPEG Compression}} & \multicolumn{4}{c|}{\textbf{FF++}} & \multicolumn{4}{c|}{\textbf{CelebDF (generalisation)}} & \multicolumn{4}{c|}{\textbf{DFDC (generalisation)}} \\
& BL & CA & EA & EA+CA & BL & CA & EA & EA+CA & BL & CA & EA & EA+CA \\
\hline
10 & 0.700 & 0.861 & 0.839 & \textbf{0.866} & 0.583 & 0.521 & 0.530 & \textbf{0.574} & \textbf{0.675}   & 0.662   & 0.661   & 0.661  \\
20 & 0.765 & 0.930 & 0.931 & \textbf{0.933} & 0.586 & 0.634 & 0.632 & \textbf{0.640} & 0.666   & 0.673   & \textbf{0.699}   & 0.697  \\
30 & 0.783 & \textbf{0.965} & 0.962 & \textbf{0.965} & 0.553 & 0.671 & 0.668 & \textbf{0.677} & 0.652   & 0.697   & 0.682   & \textbf{0.701}  \\
50 & 0.839 & 0.981 & 0.980 & \textbf{0.983} & 0.590 & 0.690 & 0.711 & \textbf{0.716} & 0.645   & 0.685   & 0.694   & \textbf{0.706}  \\
80 & 0.943 & 0.990 & 0.990 & \textbf{0.992} & 0.710 & 0.763 & 0.764 & \textbf{0.779} & 0.665   & 0.682   & 0.702   & \textbf{0.719} \\
\hline
Average & 0.806 & 0.945 & 0.940 & \textbf{0.947} & 0.604 & 0.655 & 0.661 & \textbf{0.677} & 0.660 & 0.679 & 0.687 & \textbf{0.697} \\
\hline
\end{tabular}
}
\end{table}

Table \ref{table:deepfake_jpeg} displays the AUC results of deepfake detection under different JPEG compression levels. Overall, the use of autoencoder ensembles leads to a significant improvement in performance, mainly because the encoder-decoder setup of the autoencoders operates in a manner similar to that of the JPEG compression-decompression process. The EA approach can assist in the training of the model by providing samples that are comparable to those acquired using JPEG compression, and thus can simulate the behavior of the attack. Training on data with similar behavior to the one encountered during the attack is likely to enhance the detector's performance, which is evident from the results presented in our study. 

The EA and EA+CA models outperform the BL model in all tests, exhibiting an average improvement in performance of 9.4\% and 12.1\%, respectively, on the CelebDF dataset, and 4.1\% and 5.6\% on the DFDC preview dataset. Compared to the CA model, the EA and EA+CA models exhibit an average improvement in performance of 0.9\% and 3.4\%, respectively, on the CelebDF dataset, and 1.2\% and 2.7\% on the DFDC preview dataset. This reinforces the importance of data augmentation in real-world applications, irrespective of type. 
 
As expected, the impact of strong compression on the deepfake detectors is more pronounced due to the prominent artifacts introduced by the very lossy compression. The most significant performance improvements on unknown datasets are observed for compression levels that are slightly less aggressive, as they create small artifacts that are not visible to the naked eye and are commonly encountered on the internet.

\textbf{DeepFake attacks:}
Consistent with previous research, we ascertain that forensic classifiers are exceedingly vulnerable to such adversarial attacks. Through our white-box attacks, we observe a significant decline in the AUC, decreasing from 0.9x to below 0.1x. This reduction is in stark contrast to the AUC value of 0.5, which represents a classifier that randomly guesses between "real" and "fake" without any discriminatory capability. In contrast, when we are not able to directly access the classifier’s parameters, our black-box attacks still reduce the ROC by at least 5\% on all datasets.

\begin{table}[htbp]
\centering
\caption{Deepfake detection results under blackbox and whitebox attacks expressed in AUC on the FF++, CelebDF and DFDC preview datasets where: BL--baseline model, CA--classic augmentation model, EA--ensemble autoencoder augmentation model (proposed), EA+CA--ensemble autoencoder augmentation model combined with classical augmentations (proposed).}
\label{table:deepfake_adversarial}
\renewcommand{\arraystretch}{1.2}
\resizebox{\linewidth}{!}{%
\begin{tabular}{|l|cccc|cccc|cccc|}
\hline
\multirow{2}{*}{\textbf{Adversarial Attacks}} & \multicolumn{4}{c|}{\textbf{FF++}} & \multicolumn{4}{c|}{\textbf{CelebDF (generalisation)}} & \multicolumn{4}{c|}{\textbf{DFDC preview (generalisation)}} \\
& BL & CA & EA & EA+CA & BL & CA & EA & EA+CA & BL & CA & EA & EA+CA \\
\hline
Black-box & 0.887 & 0.898 & 0.911 & \textbf{0.926} & 0.645 & 0.712 & 0.681 & \textbf{0.717} & 0.597   & 0.635   & 0.643   & \textbf{0.660}  \\
White-box & \textbf{0.014} & 0.003 & 0.009 & 0.011 & 0.001 & 0.014 & 0.027 & \textbf{0.029} & 0.010   & 0.073   & 0.069   & \textbf{0.078}  \\
\hline
\end{tabular}
}
\end{table}

For the black-box attacks, our algorithm's resistance is proportional to the performance of the model used. In this case, we used the standard ResNet-50 model to generate the attacks. We found that the proposed EA and EA+CA methods outperforms the BL method on the FF++ dataset by 2.7\% and 4.4\%, respectively. Similarly, on the CelebDF dataset, the methods exhibits a performance improvement of 5.6\% and 11.1\%, while on the DFDC preview dataset, the methods demonstrates a performance enhancement of 7.7\% and 10.6\% compared to BL. Conversely, whitebox attacks were found to fully impact the model's output. In this case, all results were below 10\% AUC, indicating that the model would make the opposite prediction over 90\% of the time. While the improvements were not substantial, our model still demonstrated a performance increase over the baselines.

\section{Conclusions}
\label{sec:conclusions}

This study focused on evaluating the effectiveness of our autoencoder ensemble-based data augmentation algorithm for deepfake detection in real-world applications. To this end, we employed basic, state-of-the-art models, prioritizing the impact of the augmentation algorithm rather than the model architecture itself. Our models were trained with generalisation in mind to ensure model-agnosticism. Additionally, the autoencoders were trained to realistically replicate images from the original dataset, by exposing the model to new deep learning fingerprints that are imperceptible to the human eye. Through our evaluation, we aimed to determine whether the proposed data augmentation algorithm can enhance the generalisation of the models to other datasets, increase their resistance against basic data perturbations, improve their resilience to lossy compression algorithms such as JPEG, and enhance their immunity to adversarial attacks. Overall, our findings indicate that the proposed autoencoder ensemble-based data augmentation algorithm can significantly improve the performance of deepfake detection models in real-world scenarios.

\section*{Acknowldgements}
Liviu-Daniel Ștefan and Bogdan Ionescu's work has been funded by the Ministry of Research, Innovation, and Digitization UEFISCDI, project DeepVisionRomania, “Identifying People in Video Streams using Silhouette Biometrics”, Solutions Axis, grant 28SOL/2021, PN-III-P2-2.1-SOL-2021-0114. 
Mihai Gabriel Constantin's work has been funded by the Ministry of Research, Innovation, and Digitization UEFISCDI, project PIMS-IAT, “Platform Specialized in Identifying and Evaluating Early Warning Indices for Crisis Management”, Solutions Axis, grant 27SOL/2021, PN-III-P2-2.1-SOL-2021-0063.
Mihai Dogariu and Andrei Jitaru's work has been funded by the Ministry of Investments and European Projects through the Human Capital Sectoral Operational Program 2014-2020, Contract no. 62461/ 03.06.2022, SMIS code 153735.
\printbibliography
\end{document}